\documentclass[runningheads]{llncs}

\usepackage{graphicx}
\usepackage[bookmarks=false]{hyperref}
\usepackage{subfig}
\usepackage{amssymb,amsmath}
\usepackage[capitalise]{cleveref}
\usepackage[utf8]{inputenc}
\usepackage{algorithm}
\usepackage{algpseudocode}
\usepackage{array}
\newcolumntype{C}[1]{>{\centering\arraybackslash}m{#1}}

\usepackage{amssymb}
\usepackage{textcomp}

\begin{document}
\title{Oil and Gas Reservoirs Parameters Analysis Using Mixed Learning of Bayesian Networks}
\titlerunning{Reservoirs Parameters Analysis using Bayesian Networks}
%
%
\author{Irina Deeva\inst{1} \and
Anna Bubnova\inst{1} \and
Petr Andriushchenko\inst{1} \and
Anton Voskresenskiy\inst{1,2} \and
Nikita Bukhanov\inst{1,3,}\thanks{- correspondence author (bnikita.itmo@gmail.com)} \and
Nikolay O. Nikitin\inst{1} \and
Anna V. Kalyuzhnaya \inst{1}}
\authorrunning{I. Deeva et al.}
%
\institute{ITMO University, Saint-Petersburg, Russia \and
Gazpromneft-GEO, Saint-Petersburg, Russia \and
Gazpromneft Science and Technology Center, Saint-Petersburg, Russia 
}
\maketitle              

\begin{abstract}

 In this paper, a multipurpose Bayesian-based method for data analysis, causal inference and prediction in the sphere of oil and gas reservoir development is considered. This allows analysing parameters of a reservoir, discovery dependencies among parameters (including cause and effects relations), checking for anomalies, prediction of expected values of missing parameters, looking for the closest analogues, and much more. The method is based on extended algorithm MixLearn@BN for structural learning of Bayesian networks. Key ideas of MixLearn@BN are following: (1) learning the network structure on homogeneous data subsets, (2) assigning a part of the structure by an expert, and  (3) learning the distribution parameters on mixed data (discrete and continuous). Homogeneous data subsets are identified as various groups of reservoirs with similar features (analogues), where similarity measure may be based on several types of distances. The aim of the described technique of Bayesian network learning is to improve the quality of predictions and causal inference on such networks. Experimental studies prove that the suggested method gives a significant advantage in missing values prediction and anomalies detection accuracy. Moreover, the method was applied to the database of more than a thousand petroleum reservoirs across the globe and allowed to discover novel insights in geological parameters relationships.

\keywords{Bayesian networks  \and Structural learning \and Causal inference \and Missing values prediction \and Oil and gas reservoirs \and Similarity detection}
\end{abstract}

\section{Introduction}
The problem of choosing oil and gas reservoir development strategy is one of the most crucial decisions made at the early stages of reservoir development by every oil and gas producer company. Almost all decisions related to fluid production have to be made at the early stages, characterized by high uncertainty and lack of information about geological and production reservoir parameters. The reservoir could be fully characterized only at mature development stages. The experience of commissioning new reservoirs shows that most of the project decisions made at the early stages of the reservoir development have a crucial impact on the development strategy and the entire project's economic feasibility. A common method of investigating these new reservoirs is to examine a subsample that is close to or similar to them. Most often, the wrong selection of analogues leads to the fact that the reserves of the reservoir are overestimated, which leads to a discrepancy between the forecast and actual production levels, which results in an overestimated net present value (NPV) forecast for the entire project \cite{jager2012influence}. Incorrect selection of analogues can even lead to the fact that the actual NPV does not fall into the predicted distribution of probable NPV. 

In a significant part of geological companies, analogues search is performed by an expert who manually selects a reservoir which properties resemble properties of the target one. The result of this procedure is a list of reservoir names with similar properties. Analogues are also used at mature stages of reservoir development, for example, to find successful cases of increasing oil recovery. There is another way to find reservoir analogues, namely using the similarity function \cite{perez2014identification}. The difference between the two approaches consists mainly in distributions shapes of reservoir analogues due to more narrow search space made by industry expert  \cite{voskresenskiydistr2020variations}. It was also founded that using a manual approach, some experts limit themselves only to local analogues and completely ignore global analogues. A decision of using only local analogues may not be optimal, especially at the early stages of project development \cite{sun2002geological}. On the other hand, the results of reservoir analogues performed by ranked similarity function are characterized by a broader distribution of reservoir parameters. In order to find reservoir analogues by similarity function more feasible, reconstruction of missing values should be performed. 
Based on this, it is highly desirable to have a more automated and mathematically proved instrument to determine the most likely reservoir parameters at the early stages of geological exploration. So far, a few works have been done on topics related to inputting missing values using machine learning to reservoir parameters datasets \cite{voskresenskiy2020feature}. For this reason, our efforts also addressed the search for an automated solution for the analysis and modelling of reservoir properties with the usage of machine learning techniques.


\section{Related Work}
In this section, we summarize, in abbreviated form, the pros and cons of methods that are used for modelling and could potentially be the basis for reservoir analysis. They are described in more detail in our previous work \cite{andriushchenko2020analysis}.

One common way of modelling is Markov Chain Monte Carlo (MCMC) method. The algorithm walks through the space of all possible combinations of values and moves from one state to a new state that differs in the $i$-th variable with probability estimated from the dataset. However, this approach for high-dimensional problems requires large amounts of memory and time. Unfortunately, this model does not find the relationship between the parameters explicitly and is difficult to interpret.  Nevertheless, MCMC is quite common, ready-made libraries exist for it, and this method is used to solve practical problems in oil and gas data. For example, Gallagher et al. \cite{gallagher2009markov} showed the applicability of MCMC method to the task of modelling distributions of geochronological ages, sea-level, and sedimentation histories from two-dimensional stratigraphic sections.

Another modelling approach is the Conditional Iterative Proportional Fitting (CIPF). CIPF is designed to work with multi-dimensional contingency tables. The basic idea is to iteratively fit marginal and conditional distributions, gradually approaching the desired joint distribution in terms of Kullback-Leibler distance. However, this requires knowledge of the dependency structure, but even so, the method converges extremely slowly. There is a simplified and faster version called IPF, which assumes a toy model with an independent set of variables. Li and Deutsch \cite{li2012implementation}, for example, use it to estimate and model facies and rock types. On the positive side, this approach is reduced to the sequential solution of linear equations, and there is a ready-made library for it.

A third approach is to use a copula. Formally, this is a multivariate distribution on an n-dimensional unit cube obtained using inverse distribution functions for some fixed family of distributions. Based on Sklar's theorem, modelling any joint distribution is reduced to approximation by copulas. For example, Han et al. \cite{han2019mathematical} use this approach to determine the tectonic settings. And in the work of Hernández-Maldonado et al. \cite{hernandez2014multivariate} they model complex dependencies of petrophysical properties such as porosity, permeability, etc. Unfortunately, expert selection of the copula family is required. The basic method takes into account only the pairwise dependence; more complex models require knowledge of the dependence structure. It must be considered that the main application of copulas is to work with quantitative values.

In short, our work seeks to circumvent the above limitations on the volume and type of data. Our goal is an interpretive approach that will allow us to combine the tools needed for oil and gas reservoir analysis on a single base. To do this, we turn to Bayesian networks, taking into account the specifics of the domain and the techniques adopted for analysis. 

\section{Problem statement}


Let us temporarily step aside and look at this issue from the side of statistics, not geology. The first thing that is important to us when we analyze data is to present it in the most effective way that allows us to extract as much hidden information as possible. 
And oil and gas reservoirs analysis impose some limitations: we need to work with mixed data, i.e. continuous, categorical, and discrete data. The objects under study have a large number of features.

The relationships of the parameters are not always linear and not always obvious, even to a specialist, but we need to be able to identify them for further interpretation. We also would like to have a possibility of taking into account the expert's opinion of the cause and effect links. There are additional restrictions on the time of the algorithm, as well as the need for certain functionality such as filling gaps and finding anomalies. So here is an additional challenge of making an optimal choice among the existing modelling approaches. Based on all of the above we focus our efforts on developing a tool that is able to (1) be interpretable and include some portion of expert knowledge (in the way of composite AI), (2) be multipurpose as a core for partial algorithms (analogues search, gaps filling, probabilistic inference, etc.), (3) be an efficiently computed. 

As the most promising basis, we choose Bayesian networks (BN). From a graphical point of view, it is a directed acyclic graph (DAG), any vertex of which represents some characteristic of the object. This structure also stores information for the vertices about the value and conditional distribution of the corresponding characteristic.
 Let $Pa_{X_i}^G$ denote the parents of the vertex $X_i$ in the structure $G$, ${NonDescendants}_{X_i}$ denote the vertices in the graph that are not descendants of the vertex $X_i$ \cite{koller2009probabilistic}. Then, for each vertex $X_i$:
\begin{equation}
({X_i} \perp {NonDescendants}_{X_i} | Pa_{X_i}^G)
\end{equation}
Then the multivariate distribution of P in the same space is factorized according to the structure of the graph G, if P can be represented as:
\begin{equation}
P(X_1,...,X_n) = \prod\limits_{i = 1}^n{P({X_i}|Pa_{X_i}^G)}
\end{equation}
However, to use Bayesian networks, it is necessary to solve two problems: learning the network structure and distributions parameters in the network nodes. Unfortunately, learning the graph structure is a complex and resource-intensive task. The number of DAGs grows super-exponentially with the number of vertices \cite{robinson1973counting}. However, there are several approaches \cite{scutari2019learns} to solving this problem, which we will discuss below. 

    \begin{figure}[ht]
    \centerline{\includegraphics[width=0.95\linewidth]{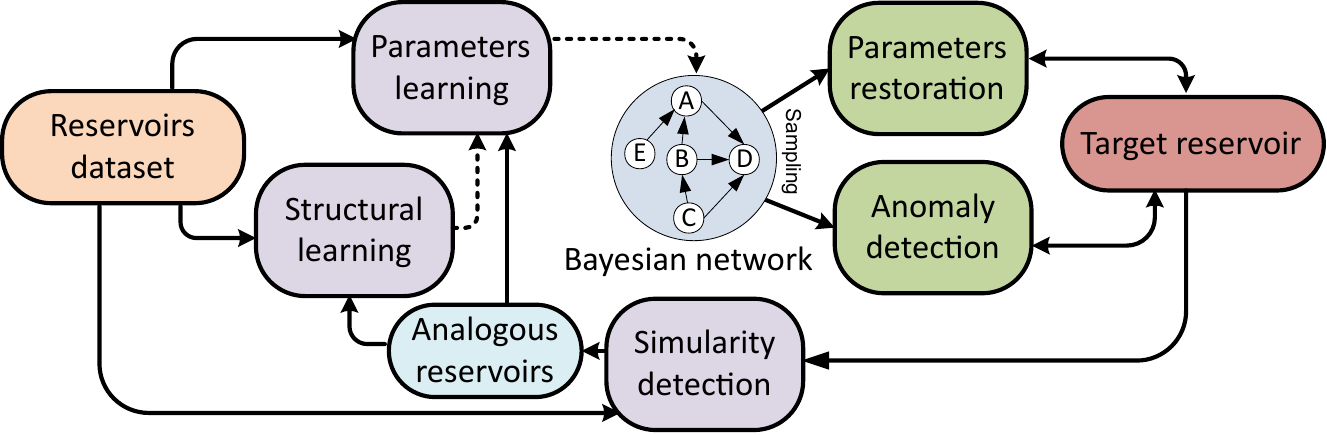}}
    \caption{The pipeline of the proposed Bayesian approach to the oil and gas reservoir analysis. Data preprocessing block are not represented directly to simplify the scheme.} 
    \label{fig_concept}
    \end{figure}

The purpose of this paper is to demonstrate a new flexible approach in the context of a comprehensive analysis of oil and gas reservoirs (see \cref{fig_concept} for details). We chose the Bayesian network model as the core because this probabilistic model generally meets the formulated criteria. There is an extensive theoretical basis for various probabilistic analysis problems, and examples of specialized implementations on small oil and gas datasets \cite{masoudi2015feature,martinelli2013building}. However, Bayesian networks are capable of handling larger amounts of data, and the functionality can be represented as a complex structure of complementary elements. In this paper, we propose to implement this idea on the basis of oil and gas reservoirs data, taking into account the specificity of the domain and typical auxiliary techniques such as searching for analogues.

\section{Algorithms and methods}
\subsection{MixLearn@BN: Algorithm for mixed learning of Bayesian networks }
Making the inference of reservoir parameters requires a complex algorithm with several crucial properties. The algorithm MixLearn@BN allows us to combine (1) learning the network structure on homogeneous data subsets, (2) assigning the structure by an expert, and  (3) learning the distribution parameters on mixed data (discrete and continuous). \cref{BN_alg} demonstrates the pseudocode of the proposed complex algorithm. For structure learning, we use the Hill Climbing algorithm \cite{chickering2002optimal}, \cite{gamez2011learning} with the K2 scoring function \cite{cooper1992bayesian}. To learn the parameters, we use a mixed approach, within which the distributions at the nodes can be of three types:
\begin{itemize}

\item Conditional probabilities tables (CPT), if the values at the node are discrete and its parents have a discrete distribution;

\item Gaussian distribution, if the values at the node are continuous and its parents have a continuous distribution;

\item Conditional Gaussian distribution, if the node values are continuous, and its parents have a discrete distribution and continuous one.
\end{itemize}

\begin{algorithm}
\caption{Comprehensive Bayesian Network Learning Algorithm}

\begin{algorithmic}[1]

\Procedure{Bayesian network learning}{$D,edges,remove\_edges$}  

    \State \underline{Input:} $D$ = $\left\{\begin{array}{ccc} \mathbf{x}_1,..., \mathbf{x}_n
                                     \end{array} \right\}$, edges which an expert wants 
    \State to add $edges$, boolean $remove\_edges$ that allows removing $edges$ 
     \State \underline{Output:} Bayesian network$\left\{
                                    \begin{array}{ccc}V, E, Distributions Parameters
                                    \end{array} \right\}$
     \State $discrete\_D$ = Discretization($D$)
     \State $\left\{\begin{array}{ccc}V, E
                                    \end{array} \right\}$ = HillClimbingSearch($discrete\_D$)  \Comment{Structure learning}
    \If{$remove\_edges = false$}        
        \State $\left\{\begin{array}{ccc}V, E
                                    \end{array} \right\}$ = $\left\{\begin{array}{ccc}V, E
                                    \end{array} \right\} \cup edges$
    \EndIf
    
    \State $bn\_parameters$ = Parameters learning($D, \left\{\begin{array}{ccc}V, E
                                    \end{array} \right\}$)             \Comment{Parameters learning}
\State \Return Bayesian network$\left\{
                                    \begin{array}{ccc}V, E, bn\_parameters
                                    \end{array} \right\}$
\EndProcedure

\Procedure{Parameters learning}{$D,\left\{\begin{array}{ccc}V, E
                                    \end{array} \right\}$}  

    \State \underline{Input:} $D$ = $\left\{
                                    \begin{array}{ccc}\mathbf{x}_1,...,\mathbf{x}_n
                                    \end{array} \right\}$, structure of BN $\left\{\begin{array}{ccc}V, E
                                    \end{array} \right\}$
     \State \underline{Output:} dictionary with distributions parameters for each node in BN
     \State $params$ = empty dictionary
     \For {$node$ in $BN\_structure$}
     
        \If{$node$ is discrete \textbf{and} parents($node$) are discrete}        
            \State $params[node]$ = CPT($node$, parents($node$), $D$)
        \EndIf
         \If{$node$ is continuous \textbf{and} parents($node$) are continuous} 
            \State $mean, var$ = parameters from Gaussian($node$,D)
            \State $coef$ = coefficients from BayesianLinearRegression(parents($node$),$node$, \State D)
            \State $params[node]$ = $\left\{
                                    \begin{array}{ccc}mean, var, coef
                                    \end{array} \right\}$
        \EndIf
        \If{$node$ is continuous \textbf{and} parents($node$) are continuous \textbf{and} discrete}
            \State $cont\_parents$ = parents\_continuous($node$)
            \State $disc\_parents$ = parents\_discrete($node$) 
            \State $node\_params$ = $\emptyset$ 
            \State $combinations$ = all combinations of $disc\_parents$ values
            \For {$\left\{\begin{array}{ccc}v_1,...,v_k
                                    \end{array} \right\}$ in $combinations$}
              \State $subsample$ = $\left\{\begin{array}{ccc}\mathbf{x}_i:x_{ij_1} = v_1,...,x_{ij_k} = v_k \end{array} \right\}$ 
              \State $mean, var$ = parameters from Gaussian($node$, $subsample$)
              \State $coef$ = coefficients from BayesianLinearRegression($cont\_parents$ , \State $node$, 
               $subsample$)
              \State $node\_params \cup \left\{
                                    \begin{array}{ccc}mean, var, coef
                                    \end{array} \right\}$
           \EndFor
           \State $params[node]$ = $node\_params$
        \EndIf

   \EndFor
\State \Return $params$
\EndProcedure

\label{BN_alg}
\end{algorithmic}
\end{algorithm}

\subsection{Reducing the training samples using similarity detection}\label{dist_metrics}

In this paper, in particular, we test whether the technique for finding similar reservoirs (analogues) can improve parameter restoration quality with Bayesian networks. The main task is to find these analogues. And for this, we need to define an appropriate measure of proximity or distance to the target reservoir.The main requirement for distance metrics are defined by task features:

\begin{enumerate} 
	\item The data is represented by a set of values of different types. Metrics on categorical variables rely on whether or not the value matched, whereas, for quantitative ones, it matters how much it does not match. Once quantitative variables are discretized, we can use the general approach for categorical ones, but there is a risk that we miss valuable information.
	\item Variables are interdependent, and mismatches cannot be considered as unrelated in any way. It is unclear how to account for this.
	\item Some variables are more valuable than others. There is a weighted option for most distances and events, but it is not obvious how to select optimal weights.
\end{enumerate} 

The following is a description of the distances between the objects $u$ and $t$ involved in the experiments. We begin by determining Gower's general similarity coefficient $S(u,t)$ \cite{gower1971general}. A auxiliary coefficient $S_j(u,t)$ is considered for each $j$th variable. And $S(u,t)$ is their weighted average with weights $w_j$. In the unweighted version, $w_j=1$ is assumed. On categorical variables, $S_j(u,t)$ is 0 or 1, depending on whether the categories match. On quantitative variables, it is the modulus of the difference of the normalized values. Note that this is a similarity coefficient, not a distance. However, it is easy to turn it into distance with the following transformation: $dist_G(u,t)=1-S(u,t).$

There are also measures of a different nature than Gower's coefficient. This is cosine distance that has proven itself in the task of ranking search engine results. Applying this distance requires prior preparation of values. For a categorical variable, the value is assumed to be 1 at the target. And 0 or 1 on the object being compared, depending on whether the categories match. For quantitative variables, values are normalized.

We also investigated the performance of the filtering function in the experiments. It depends on only one parameter $\varepsilon$, which for quantitative variables says that a value is close if $|u_j-t_j|\leq\varepsilon\cdot range(j)$. And for categorical, it checks for category matching. 
First, the set of analogues includes objects that are close to the target in all variables. Then for all but one, and so on.

\subsection{Method for parameters restoration with Bayesian networks learning on analogues}
\label{Method_restor}
Using similarity metrics allows you to find the closest reservoirs to a target reservoir. Combining the search for analogous reservoirs and learning Bayesian networks with MixLearn@BN allows formulating a method for solving the problem of parameters restoration (\cref{fig_concept}), the structure of which implies the following steps:
\begin{enumerate} 
\item Select the target reservoir from a dataset;
\item Looking for N nearest reservoirs according to the distance metric;
\item Learn the structure and parameters at the nearest reservoirs with algorithm MixLearn@BN;
\item Initialize the nodes of the Bayesian network with the values of those parameters that are not missing;
\item Sample the missing values from this Bayesian network by forward sampling with evidence \cite{henrion1988propagating};
\item For categorical values, the gap is filled with the most frequent category in the sample;
\item For continuous values the gap is filled with the average value in the sample.
\end{enumerate}

\section{Experiments and Results}
    \subsection{Exploratory data analysis with Bayesian networks}

    The dataset used in the study was collected from open sources and presented by plain database contains 1073 carbonate and clastic reservoirs from all over the world. The parameters that define the dataset contain categorical and continuous values: reservoir depth and period, depositional system and environment, tectonic regime, structural setting and trapping mechanism, lithology type, gross and net thicknesses, reservoir porosity and permeability.

    \begin{figure}[h]
    \centerline{\includegraphics[width=1\linewidth]{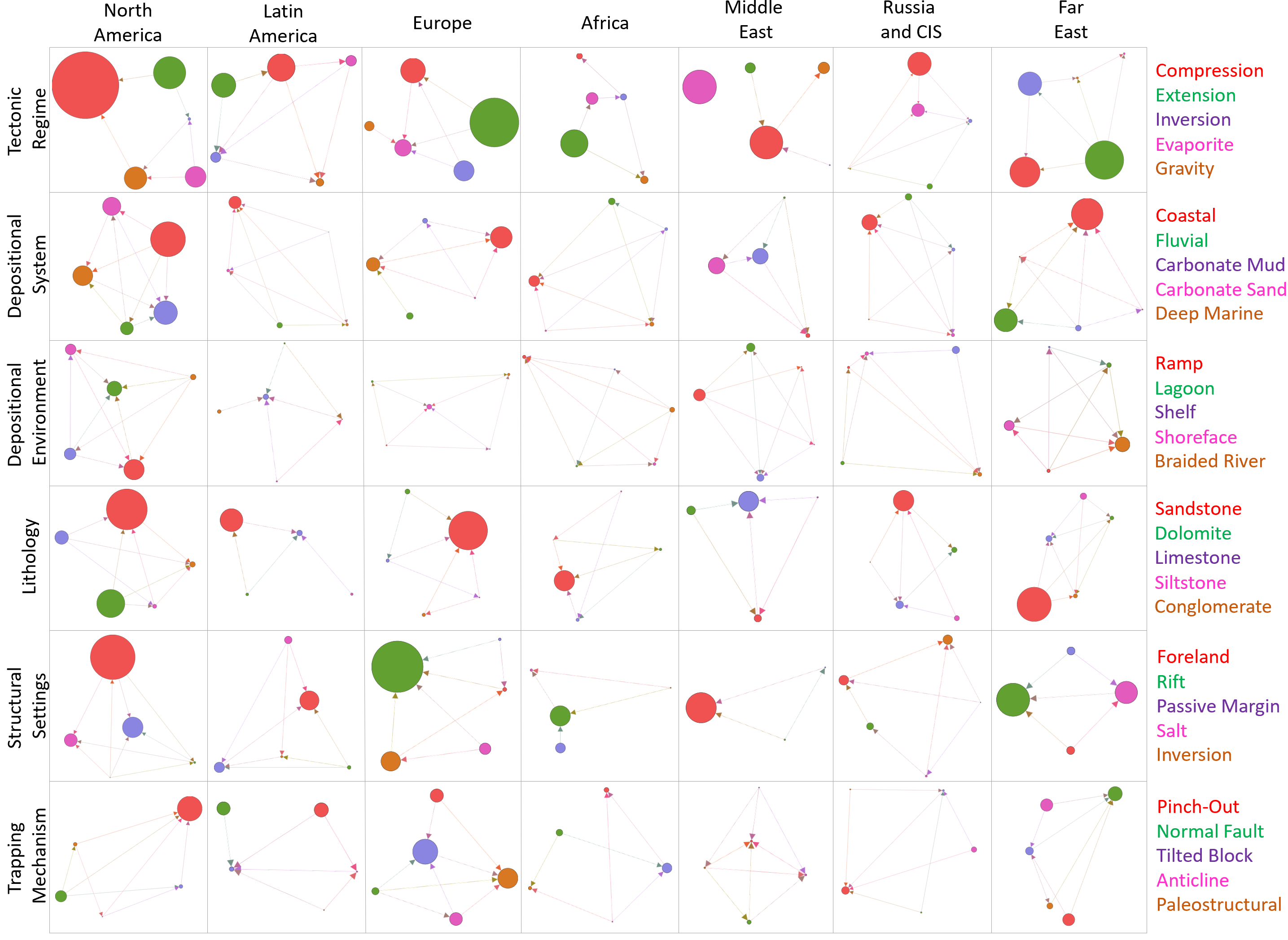}}
    \caption{Bayesian networks are for six main geological parameters across different regions worldwide. Each network has five nodes, which are described on the right margin.}
    \label{fig:variables}
    \end{figure}
    
     Several Bayesian networks were built to demonstrate that their output could be used to get qualitative data insights. The x-axis on \cref{fig:variables} signifies region division based on geographical closeness (CIS stands for Commonwealth of Independent States) and can be considered as filters of the dataset. The y-axis indicates some categorical parameters which characterize reservoirs in the dataset. The network nodes' color and size signify the value of a categorical parameter (presented on the right portion of the figure) and several such occurrences, respectively. Generally, the figure demonstrates relationships between selected reservoir parameters within regions.

    For instance, from \cref{fig:variables} following conclusion could be made: in Europe and Africa, extension tectonic regime and rift structural setting prevail. Extension and rift are closely related in terms of tectonic. Extension tectonic regime causes extension of continental lithosphere which carriers origin of rift structural setting, so the area undergoes extensional deformation (stretching) by the formation and activity of normal faults. The East African Rift System and the North Sea rift could be examples of extensional regimes in the regions. Also, there are arrows from normal faults to tilted blocks that could indicate causal inference between these parameters, and if so, this is in agreement with geological knowledge. A similar approach (in terms of analyzing statistical model with and domain knowledge) was performed by analyzing the causal inference between reservoir parameters and well log data \cite{voskresenskiy2020feature}. They have found that those statistical models could produce conclusions that mimic interpretation rules from a domain knowledge point of view, which opens up an opportunity to reveal causal relations between features. In general, such plots could be used for qualitative analysis of prevailing parameters or patterns within regions as an alternative to conventional screening performed by a geologist using manual filtering and spatial visualization techniques. 
    
     \begin{figure}[h]
    \centerline{\includegraphics[width=0.95\linewidth]{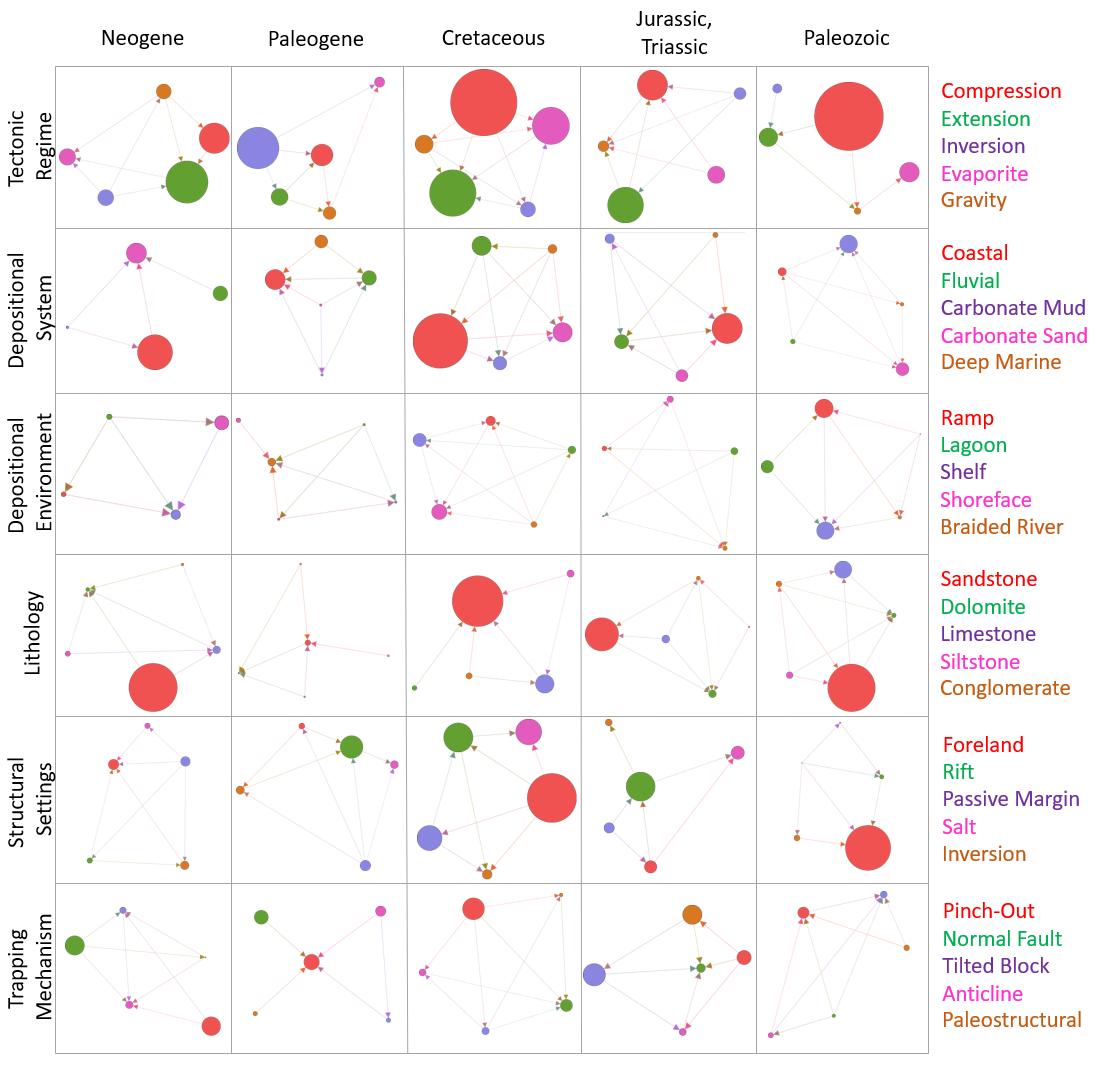}}
    \caption{Bayesian networks are for six main geological parameters across different stratigraphy periods. Each network has five nodes, which are described on the right margin.} 
    \label{fig:strata}
    \end{figure}
    
    The x-axis on \cref{fig:strata} signifies the geological period used to filter the dataset and build networks. The figure shows a relation of reservoir parameters within geological timeframes according to International Chronostratigraphic Chart \cite{cohen2013ics}. The Neogene, Cretaceous, Jurassic, and Triassic are characterized by the fact that coastal and fluvial depositional systems and sandstone lithology dominate in these geological periods. It does not contradict domain knowledge that sandstone tends to dominate in coastal and fluvial depositional systems as it the characterized by relatively high flow energy. In Paleozoic, sandstone lithology also prevalent, but carbonate mud depositional system is predominant. The reason for this discrepancy may be due to missing values in the depositional system. It is difficult to draw conclusions from the figure and confirm them with domain knowledge (as was done for the previous one) because different portions of the planet at the same geological time undergo various geological processes. 

    The possible implementation of this workflow can be easily extended to the internal company database. Apart from the fact that it has a similar number of parameters, it has many instances subdivided in hierarchical order into reservoirs, formations, and wells, respectively. This allows us to work internally with the same approach using an internal database, but here we present some results using the dataset obtained from public sources.

\subsection{Bayesian networks application experiments}\label{experiments}   
\begin{figure}[ht]
    \centerline{\includegraphics[width=0.6\linewidth]{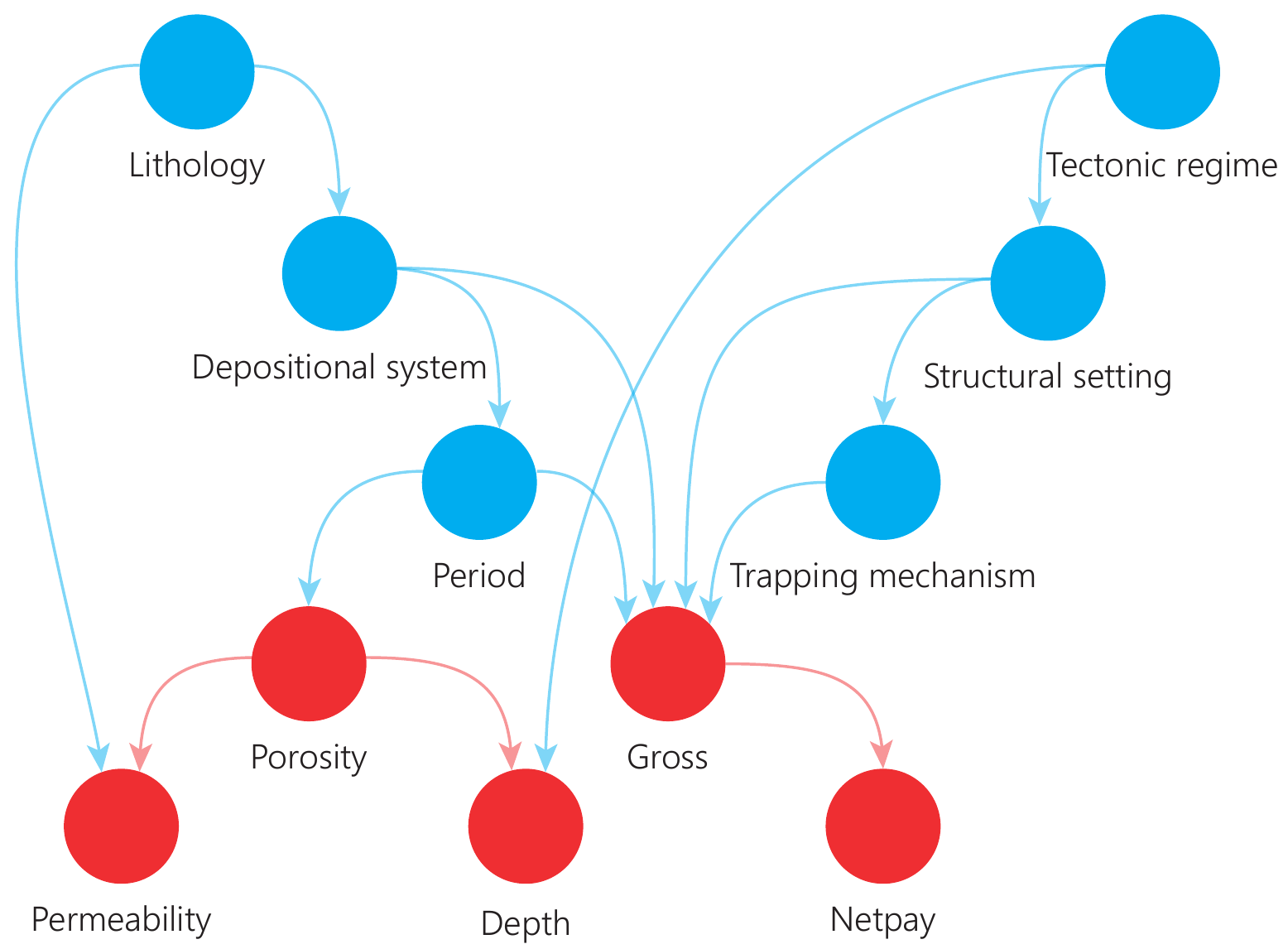}}
    \caption{The result of structural learning of the Bayesian network on the parameters of the reservoir data. The red nodes represent continuous variables and blue nodes represent the categorical variables.} 
    \label{fig:network1}
    \end{figure}

In this section, we carried out several experiments to study the Bayesian network's ability to restore missing values in parameters and detect abnormal values and compare various metrics for analogues searching in terms of the restoration accuracy. Firstly a Bayesian network was learned on the selected parameters of all reservoirs in the dataset. For structure learning, quantile data discretization was performed. The resulting Bayesian network is shown in \cref{fig:network1}. The structure of the experiment is as follows:
\begin{enumerate}
    
\item A reservoir is selected from the dataset;
    
\item The Bayesian network is learned from all reservoirs except the selected one;
    
\item In the selected reservoir, the parameters are deleted and restored with the Bayesian network;

\item Recovery results are saved;
   
\item The steps are repeated for the next reservoirs in the dataset.
 \end{enumerate}
   
To increase parameter restoration accuracy, we can learn Bayesian networks only on similar reservoirs because analogous reservoirs are more homogeneous subsamples. Our study uses several distance metrics to search for similar reservoirs (\cref{dist_metrics}). We apply method (\cref{Method_restor}) with 40 nearest reservoirs to the selected reservoir with different distance metrics. The number of nearby reservoirs was selected on the assumption that it cannot be too small to avoid overfitting, but it cannot be too large to prevent dissimilar reservoirs from entering the list of analogues.

\begin{figure}[ht!]
    \centering
    \subfloat[\label{fig:cut_penalty}]{\includegraphics[width=0.45\textwidth]{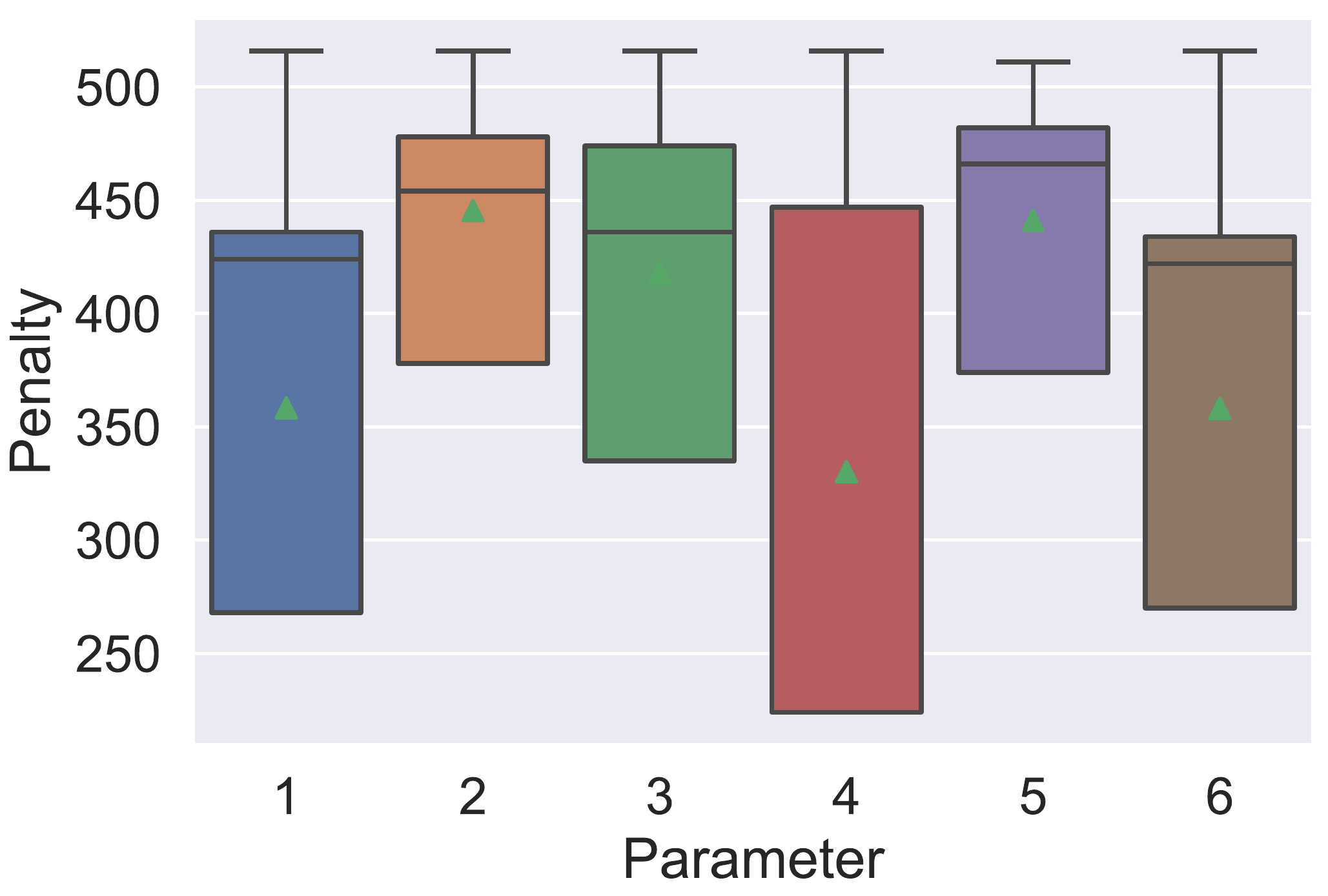}}
    \subfloat[\label{fig:num_penalty}]{\includegraphics[width=0.45\textwidth]{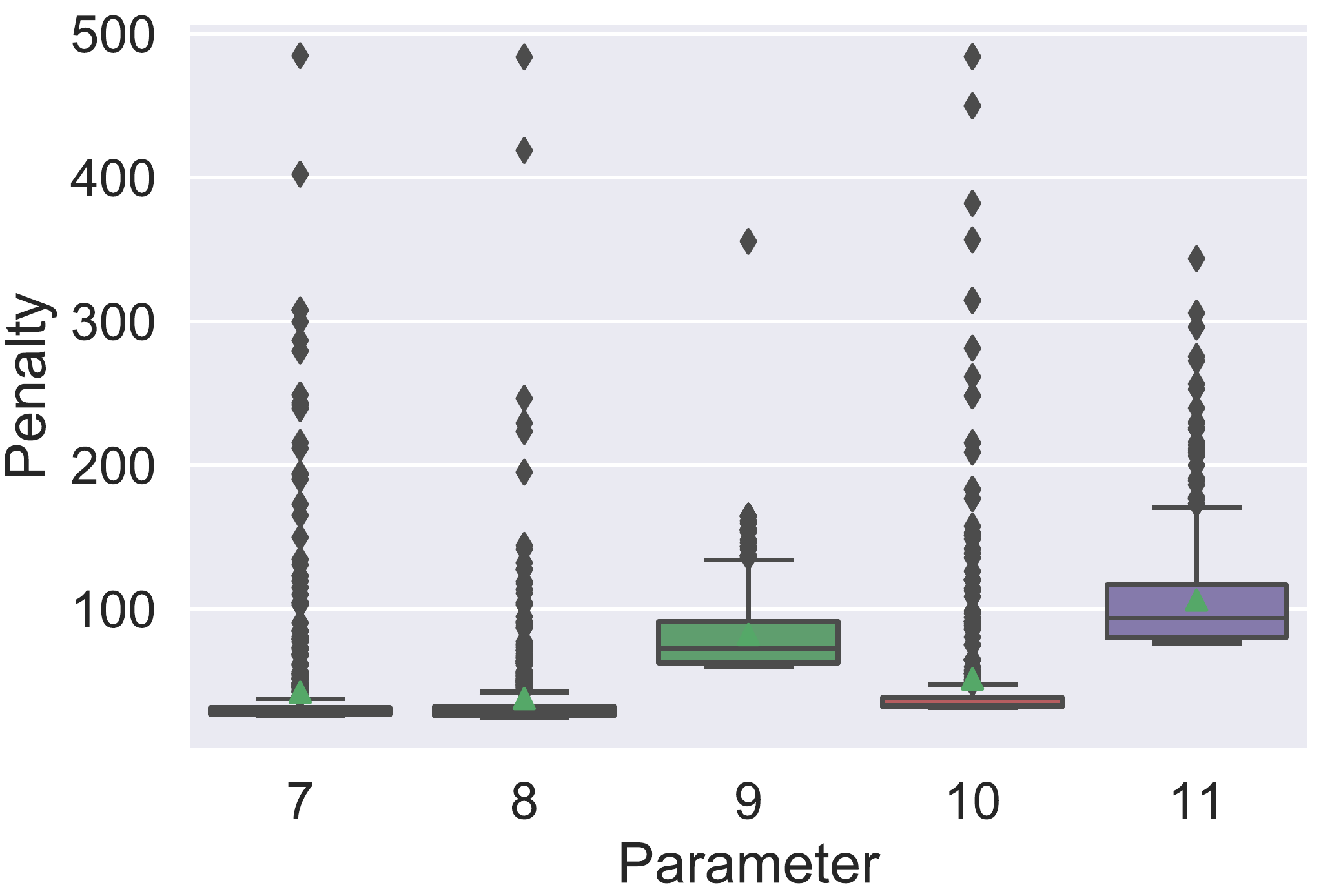}}
	\caption{The spread of penalties for categorical parameters (Tectonic regime, Period, Depositional system, Lithology, Structural setting, Trapping mechanism) (a) and continuous parameters (Gross, Netpay, Porosity, Permeability, Depth) (b) for the Gower distance, taken for all reservoirs from the dataset.}
	\label{fig:pen}
\end{figure}

Restoration results presented in \cref{table_restoration_results}. We should pay particular attention to the tables' last columns, which offer results for subsamples built at the Gower distance with weights. The point is that the Gower distance penalty for non-coincidence of categorical is greater than for non-coincidence of continuous ones. Therefore, if you equalize the penalties for all parameters, you can increase the prediction of continuous parameter values while maintaining a good prediction accuracy for categorical values. We first analyzed the average Gower distance penalties for all parameters; the spread of penalties is shown in \cref{fig:pen}. Then, to equalize the penalties, we assign the weights of the continuous parameters equal to the average penalty ratio for categorical to the average penalty for continuous ones. This means that categorical parameters are taken without weight (equal to 1), and continuous parameters are taken with a selected weight (5.8). An increase in the continuous ones' accuracy indicates that the selected value of the weights equalizes the penalties and allows you to accurately search for similar reservoirs both in terms of categorical and continuous parameters.

\begin{table}[!ht]
\caption{The values of accuracy score for categorical parameters restoration and RMSE for continuous parameters restoration.}
\label{table_restoration_results}
\begin{center}
    \begin{tabular}{|c|c|c|c|c|c|}
\hline
Parameter           & All dataset & \begin{tabular}[c]{@{}c@{}}Cosine\\ distance\end{tabular} & \begin{tabular}[c]{@{}c@{}}Gower\\ distance\end{tabular} & \begin{tabular}[c]{@{}c@{}}Filtering\\ function\end{tabular} & \begin{tabular}[c]{@{}c@{}}Gower distance\\ with weights\end{tabular} \\ \hline
\multicolumn{6}{|c|}{Accuracy for the categorical parameters}                                                                                                                                                                                                                                   \\ \hline
Tectonic regime     & 0.48        & 0.85                                                      & 0.85                                                     & \textbf{0.9}                                                 & 0.78                                                                  \\ \hline
Period              & 0.36        & \textbf{0.65}                                             & 0.63                                                     & 0.62                                                         & 0.62                                                                  \\ \hline
Depositional system & 0.56        & 0.81                                                      & 0.78                                                     & 0.76                                                         & 0.72                                                                  \\ \hline
Lithology           & 0.57        & 0.8                                                       & \textbf{0.81}                                            & \textbf{0.81}                                                & \textbf{0.81}                                                         \\ \hline
Structural setting  & 0.56        & 0.72                                                      & \textbf{0.73}                                            & 0.71                                                         & 0.71                                                                  \\ \hline
Trapping mechanism  & 0.51        & 0.76                                                      & \textbf{0.77}                                            & 0.75                                                         & 0.77                                                                  \\ \hline
\multicolumn{6}{|c|}{RMSE for the continious parameters}                                                                                                                                                                                                                                        \\ \hline
Gross               & 399.92      & 416.59                                                    & 384.98                                                   & 375.37                                                       & \textbf{306.61}                                                       \\ \hline
Netpay              & 89.7        & 94.69                                                     & 77.65                                                    & 75.84                                                        & \textbf{68.66}                                                        \\ \hline
Porosity            & 6.09        & 7.04                                                      & 6.23                                                     & 6.24                                                         & \textbf{4.62}                                                         \\ \hline
Permeability        & 1886.06     & 1450.2                                                    & 1359.97                                                  & 1271.64                                                      & \textbf{846.01}                                                       \\ \hline
Depth               & 1372.47     & 1088.82                                                   & 1052.4                                                   & 1126.7                                                       & \textbf{779.14}                                                       \\ \hline
\end{tabular}
\end {center}
\end{table}

The most challenging case of anomalies in reservoir data is when the reservoirs themselves' continuous parameters are not out of the range of possible values, but in combination with other characteristics, they are impossible. We experimented with testing the Bayesian network's ability to search for such anomalous cases. In each continuous parameter, 10\% of the values were randomly changed; such a change simulates the anomaly in the parameter. Then the nodes of the Bayesian network were initialized with other parameters, and the checked parameter was sampled. If the current value fell outside the interval of two standard deviations of the sample, it was recognized as an anomaly. ROC-AUC metrics for searching anomalous values of reservoirs parameters [Gross, Netpay, Porosity, Permeability, Depth] are equal to [0.85, 0.97, 0.8, 0.71, 0.7]. 


\section{Conclusion}
In this paper, a multipurpose method for analysis of heterogeneous data was presented using data from oil and gas reservoirs. This method consists of constructing Bayesian networks with mixed learning algorithm MixLearn@BN to enhance the accuracy. First, the learning of distribution parameters on mixed data was proposed. Thus, the accuracy of restoring continuous parameters has increased. Secondly, an approach was proposed for training the structure and parameters of a Bayesian network on a subsample of similar reservoirs. This made it possible to make the distributions of most parameters more unimodal, which led to a significant increase in restoration accuracy in all parameters at once. To find similar reservoirs, we used several distance metrics that can work with discrete and continuous data. The highest restoration accuracy for most categorical variables was obtained for the Gower distance. Using the Gower distance with weights allowed us to maintain sufficient accuracy for categorical parameters and significantly improve continuous ones' accuracy. Also, the Bayesian network showed fairly good accuracy in searching for anomalies. 

In the future, for the development of this approach, it would be interesting to consider other distances for heterogeneous data. It would also be interesting to describe other practical cases, such as quality control, studying one parameter's effect on another, and others.

\section{Acknowledgement}
We would like to thank Gazprom Neft for the provided reservoir dataset. 

%
%
%
\bibliographystyle{splncs04}
\bibliography{mybibliography}

\end{document}